\documentclass[sigconf]{acmart}
\AtBeginDocument{%
  }

\DeclareMathOperator*{\argtopk}{arg\,top-k}
\DeclareMathOperator*{\argmax}{arg\,max}

\usepackage{multirow}
\usepackage{array}
\usepackage{booktabs}
\usepackage{subcaption}
\usepackage{enumitem}
\usepackage{balance}

\copyrightyear{2026}
\acmYear{2026}
\setcopyright{cc}
\setcctype{by}
\acmConference[CIKM '26]{Proceedings of the 35th ACM International Conference on Information and Knowledge Management}{November 07--11, 2026}{Rome, Italy}
\acmBooktitle{Proceedings of the 35th ACM International Conference on Information and Knowledge Management (CIKM '26), November 07--11, 2026, Rome, Italy}
\acmDOI{10.1145/3799682.3840091}
\acmISBN{979-8-4007-2539-5/2026/11}

\begin{document}

\title[Bidding-Aware Retrieval for Multi-Stage Consistency in Online Advertising]{Bidding-Aware Retrieval for Multi-Stage Consistency in Online Advertising}

\author{Bin Liu}
\authornote{Both authors contributed equally to this research.}
\orcid{0000-0002-6044-0222}
\email{zhuoli.lb@alibaba-inc.com}
\author{Yunfei Liu}
\authornotemark[1]
\orcid{0000-0001-6377-1513}
\email{lyf327482@alibaba-inc.com}
\affiliation{
  \institution{Taobao \& Tmall Group of Alibaba}
  \city{Beijing}
  \country{China}
}

\author{Ziru Xu}
\orcid{0009-0007-4333-1212}
\email{ziru.xzr@alibaba-inc.com}
\author{Zhi Kou}
\orcid{0009-0008-0005-809X}
\email{kouzhi.kz@alibaba-inc.com}
\author{Yeqiu Yang}
\orcid{0009-0005-1717-6963}
\email{yangyeqiu.yyq@alibaba-inc.com}
\affiliation{
  \institution{Taobao \& Tmall Group of Alibaba}
  \city{Beijing}
  \country{China}
}

\author{Han Zhu}
\correspondingauthor
\orcid{0000-0002-9522-5637}
\email{zhuhan.zh@alibaba-inc.com}
\author{Jian Xu}
\orcid{0000-0003-3111-1005}
\email{xiyu.xj@alibaba-inc.com}
\affiliation{
  \institution{Taobao \& Tmall Group of Alibaba}
  \city{Beijing}
  \country{China}
}

\renewcommand{\shortauthors}{Bin Liu et al.}

\begin{abstract}

Online advertising systems typically use a cascaded architecture to manage massive requests and candidate volumes, where the ranking stages allocate traffic based on eCPM (predicted CTR × Bid). With the increasing popularity of auto-bidding strategies, the discrepancy between the computationally sensitive retrieval stage and the ranking stages becomes increasingly critical, as the former cannot access precise, real-time bids for the vast ad corpus. This misalignment inevitably leads to revenue leakage for the platform and sub-optimal ROI for advertisers.
To tackle this problem, we propose \textbf{B}idding-\textbf{A}ware \textbf{R}etrieval (\textbf{BAR}), a model-based retrieval framework that addresses multi-stage discrepancy by incorporating ad bid value into the retrieval scoring function. The core innovation is Bidding-Aware Modeling, utilizing bid signals through monotonicity-constrained learning and multi-task distillation to ensure economically coherent representations, while Asynchronous Near-Line Inference enables real-time updates to the embedding for market responsiveness. Furthermore, the Task-Attentive Refinement module selectively enhances feature interactions to disentangle user interest and commercial value signals. Extensive offline experiments and full-scale deployment across Alibaba's display advertising platform validated BAR's efficacy: 4.32\% platform revenue increase, with a 22.2 percentage-point gain in the Impression Improved Ratio for ads receiving positive operational adjustments.

\end{abstract}



\maketitle

\section{Introduction}

\begin{figure}[!tb]
  \centering
  \includegraphics[width=\linewidth]{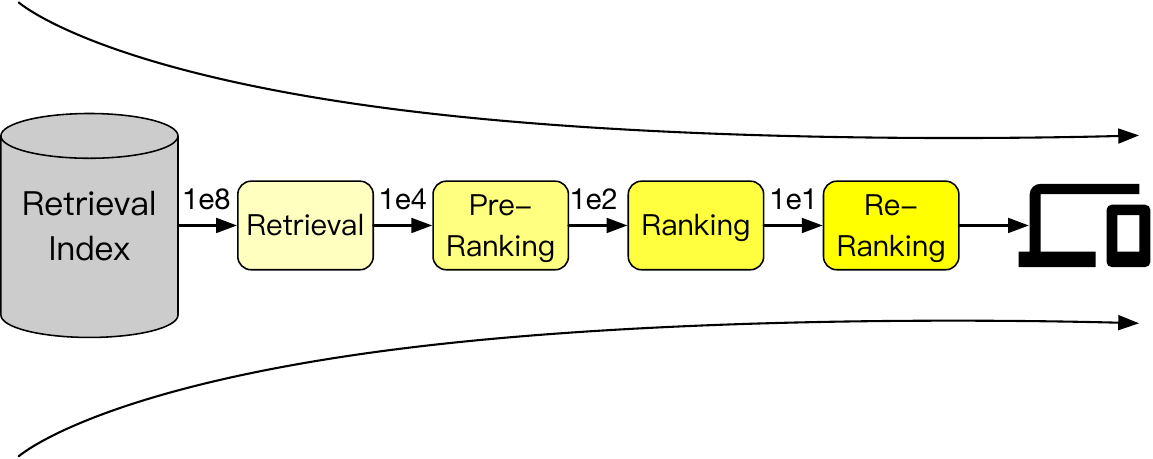}
  \caption{An illustration of the typical multi-stage cascade architecture of the online advertising platform. The numbers indicate the candidate volume at each stage.}
  \Description{A funnel diagram of the four-stage cascade: retrieval, pre-ranking, ranking, and re-ranking, with the candidate volume shrinking from billions of ads to tens of impressions.}
  \label{fig:arch}
\end{figure}

Modern industrial-scale recommender systems and online advertising platforms widely adopt multi-stage cascaded architectures~\cite{covington2016deep, 10.1145/2009916.2009934, liu2017cascade, qin2022rankflow, zheng2024full, 10.1145/3077136.3080819} to efficiently manage large-scale corpora under strict latency and throughput requirements.
As illustrated in Figure~\ref{fig:arch}, this architecture comprises Retrieval~\cite{huang2020embedding,zhu2018learning}, Pre-Ranking~\cite{Zhao_2023}, Ranking~\cite{din, dien}, and Re-Ranking~\cite{pei2019personalized}, progressively refining billion-scale candidates to the tens of impressions delivered to a user, which makes multi-stage consistency a critical determinant of overall system efficiency.

\begin{figure}[!tb]
  \centering
  \includegraphics[width=\linewidth]{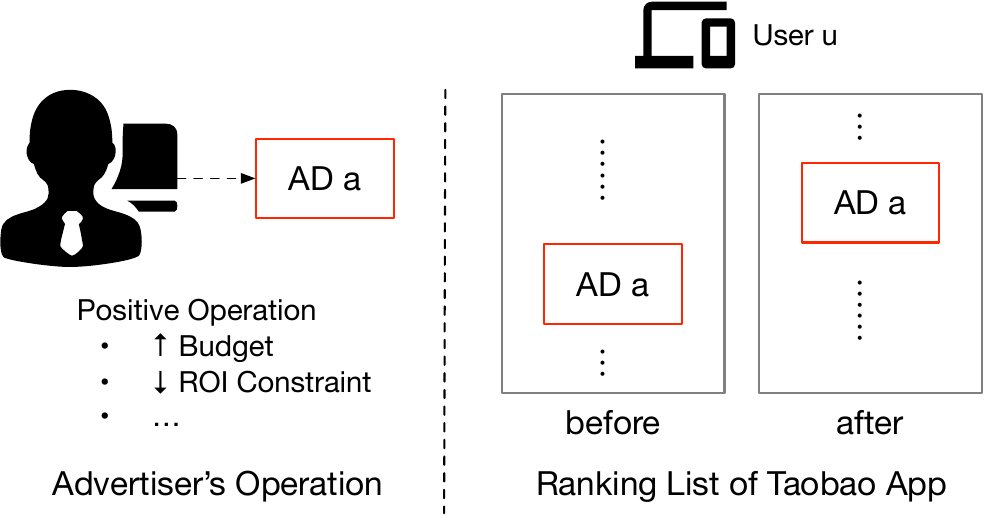}
  \caption{An illustration of the influence of ad operation.}
  \Description{A diagram showing that after an advertiser applies a positive operation such as a budget increase or an ROI constraint reduction, the bid of the corresponding ad rises and the ad moves to a higher ranking position in the auction.}
  \label{fig:ad-operation}
\end{figure}

On such platforms, which act as intermediaries between advertisers and prospective customers~\cite{evans2009online, goldfarb2011online, dehghani2015research, zhu2017optimized}, eCPM is the primary ranking criterion: advertisers compete for impressions by submitting bids, and the platform ranks ads by predicted eCPM to maximize cumulative revenue.
The ranking stages therefore model CTR prediction and bid generation for each ad campaign per auction event, and evolving automated bidding strategies~\cite{zhu2017optimized, perlich2012bid, yuan2014empirical, guo2024generative, wang2015real, 10.1145/3447548.3467199, Jin_2018, 10.1145/3219819.3219918} keep improving both bid accuracy and responsiveness to advertiser operations. As depicted in Figure~\ref{fig:ad-operation}, a positive adjustment (e.g., a budget increase or an ROI constraint reduction) raises an ad's bid and, ceteris paribus, its ranking position.

At the retrieval stage, however, running accurate CTR prediction and bid generation over billion-scale candidates is computationally prohibitive, and stronger auto-bidding algorithms~\cite{guo2024generative, 10.1145/3447548.3467199, 10.1145/3219819.3219918} only widen the resulting inconsistency with the ranking stages. Ads with high predicted bids are thus underestimated during retrieval, at the cost of platform revenue and advertiser outcomes.

To rectify this foundational discrepancy, we propose \textbf{BAR}, a \textbf{B}idding-\textbf{A}ware \textbf{R}etrieval framework. Instead of predicting the CTR and the bid of each ad separately, BAR learns to rank ads by eCPM, which remains computationally feasible over a large corpus, and makes that ranking react to bid dynamics both during training, through Bidding-Aware Modeling and a Task-Attentive Refinement module, and during serving, through an Asynchronous Near-Line Inference service.

Stage-wise consistency itself is well studied one stage downstream: pre-ranking models such as COPR~\cite{Zhao_2023} already align with the ranking stage through learning-to-rank and distillation, and we borrow that formulation rather than claim it. What has remained out of reach is the retrieval stage, where the corpus is orders of magnitude larger, per-request bid evaluation is infeasible, and ad embeddings are pre-computed, so retrieval models stay bid-agnostic. To the best of our knowledge, BAR is the \textbf{first industrial-scale retrieval framework that directly ingests dense bidding features and reacts to bid changes in near-real time}. Our primary contributions are summarized as follows:
\begin{itemize}[leftmargin=*]
  \item We propose a \textbf{Bidding-Aware Modeling} paradigm that integrates dense bidding dynamics into the retrieval stage within its efficiency budget, combining a learning-to-rank objective with a bidding-aware loss that enforces monotonicity between scores and bids and an auxiliary distillation task for representation enrichment. The accompanying Asynchronous Near-Line Inference service delivers real-time bid perception in a high-QPS production environment without performance degradation.
  \item We introduce a \textbf{Task-Attentive Refinement} module that disentangles and selectively enhances feature interactions for user interest and commercial value, preventing the marginalization of high-value ad candidates.
  \item Extensive offline experiments and full-scale deployment in Alibaba's display advertising system validate BAR's effectiveness: a 4.32\% increase in platform revenue and a 3.78\% lift in RPM at a stable CTR (+0.31\%). Furthermore, BAR empowers advertisers to actively optimize campaign outcomes, evidenced by gains of 6.6 and 22.2 percentage points in the Retrieval and Impression Improved Ratios for ads with positive operations.
\end{itemize}

\section{Related Works}
\subsection{Retrieval Modeling for Online Advertising}
Early retrieval models~\cite{zhu2018learning,zhu2019joint,zhuo2020learning, 10.1145/2505515.2505665} in online advertising optimized a specific metric such as CTR or eCPM by direct value prediction. Low-latency serving over a massive corpus, however, caps retrieval-model complexity, which opens a representational and objective gap to the far more sophisticated ranking stage~\cite{din, dien, sheng2024enhancing, Chang_2023, guo2017deepfm, pi2020search}.
The Learning-to-Rank (LTR) paradigm~\cite{qin2022rankflow,zheng2024full,10.1145/3077136.3080819, Tang2018,gallagher2019joint,huang2023cooperative} narrows that gap: standard in pre-ranking~\cite{Zhao_2023} and usually implemented as knowledge distillation from the successor stage, it trains the retriever to mimic the scoring behavior of the more powerful ranking model. One limitation survives, though: the retriever stays bid-agnostic. Trained on signals that carry bidding information only implicitly, if at all, it cannot respond to fluctuating bids and budgets, which is the gap Bidding-Aware Modeling closes.

\subsection{Retrieval Framework}
Embedding-Based Retrieval (EBR)~\cite{covington2016deep, 10.1145/2505515.2505665, muja2014scalable, shrivastava2014asymmetric} set the initial standard, using a two-tower architecture that encodes user and item features into separate dense embeddings and combines them by an inner product. Multi-interest models~\cite{li2019multi, Cen2020ControllableMF, Meng2022CoarsetoFineKM,10.1145/3604915.3608766} extended it with several embedding vectors per user to capture the many facets of user preference.
Model-Based Retrieval (MBR)~\cite{zhu2018learning,zhu2019joint, zhuo2020learning, gao2020deep, Chen2022} then replaced the inner product with richer interactions, such as cross-attention and MLP layers inside the scoring function, capturing fine-grained, context-aware relationships at substantially higher capacity. To keep full-corpus evaluation affordable, MBR pairs the scorer with an efficient index: tree structures~\cite{zhu2018learning,zhu2019joint,zhuo2020learning}, learnable paths~\cite{gao2020deep}, or graph-based indexes~\cite{Chen2022}.
We build on MBR, but observe that fusing all features into a single interaction space dilutes and entangles distinct signals; our Task-Attentive Refinement module enhances the user-interest and the commercial (i.e., bidding) signals separately instead.

\subsection{Asynchronous Near-Line Inference}
Large-scale retrieval trades serving efficiency, obtained by pre-computing target embeddings, against responsiveness. Asynchronous Near-Line Inference is the standard industrial answer, offloading computation from the online path~\cite{chandramouli2011streamrec,li2021truncation,deng2025onerec, Zhao_2023}: Streaming VQ~\cite{bin2025real} updates item embeddings and the corresponding VQ index asynchronously for index immediacy, and RTAMS-GANNS~\cite{10.1145/3627673.3680054} inserts new vectors into the ANN index in real time through multi-stream parallel execution.
We adapt the principle to static, pre-computed ad embeddings, refreshing them in near-\textbf{real-time} as market signals arrive.

\begin{table*}[!h]
\centering
\caption{Bid features used in BAR, together with the perturbation applied by the Bidding-Aware Objective and the resulting monotonicity indicator $\mathbb{I}$. Features left unperturbed carry no monotonicity constraint.}
\begin{tabular}{llll}
\toprule
\textbf{Feature} & \textbf{Description} & \textbf{Perturbation} & \textbf{$\mathbb{I}$} \\
\midrule
bid\_type & Ad bid type: OCPC~\cite{zhu2017optimized}, Max Return, or Multi-constrained bidding~\cite{guo2024generative} & --- & --- \\
bid\_constraint & KPI constraint of the campaign, such as cost per click or per conversion & relax / tighten & $+1$ / $-1$ \\
budget\_left\_ratio & Remaining budget ratio of the ad within the day, in $[0,1]$ & add / reduce budget & $+1$ / $-1$ \\
time\_left\_ratio & Remaining time ratio within the day, in $[0,1]$ & --- & --- \\
default\_bid\_value & Historical statistical mean of pBid for the ad & --- & --- \\
\bottomrule
\end{tabular}
\label{tab:bid_feature}
\end{table*}

\section{Methodology}
\subsection{Preliminary}\label{sec:preliminary}

\paragraph{Problem Definition}
Let \(\mathcal{U}\) and \(\mathcal{A}\) be the sets of user requests and ads. For a request \(u\in\mathcal{U}\), retrieval selects a size-\(k\) subset \(S\subset\mathcal{A}\) that maximizes the aggregate expected Cost Per Mille (eCPM):
\begin{equation}
\argmax_{S\subset\mathcal{A},\,|S|=k} \quad \mathrm{eCPM}(S\mid u).
\end{equation}
Direct optimization over \(\mathcal{A}\) is intractable at industrial scale, hence the cascade of Figure~\ref{fig:arch}. As its first stage, retrieval reduces to point-wise top-\(k\) scoring:
\begin{equation}
\argtopk_{a\in\mathcal{A}}\; f_{\text{eCPM}}(u, a)\,,
\end{equation}
where the learned \(f_{\text{eCPM}}(u, a)\) estimates the eCPM of ad $a$ for request $u$, and the selected ads form the candidate pool for the more expensive stages downstream.

\paragraph{Dataset}
We build BAR's training and test sets from sampled online serving logs containing both impression and ranking records:
\begin{align}
\mathcal{D}&= \{(u,\, \mathcal{I}_u \cup \mathcal{R}_u)\}, \\
\mathcal{I}_u &= \{(a_i,\, \text{eCPM}_i)_{i=1}^{|\mathcal{I}_u|}\}, \quad
\mathcal{R}_u = \{(a_i,\, \text{eCPM}_i)_{i=1}^{|\mathcal{R}_u|}\},
\end{align}
Here \(\mathcal{I}_u\) collects the ads that were finally exposed for request $u$, while \(\mathcal{R}_u\) collects those that reached the ranking stage but lost the auction, and each \(\mathrm{eCPM}_i=\mathrm{pCTR}_i\times\mathrm{pBid}_i \times 1000\) is the ranking stage's own prediction for that ad and request.
Each sample carries user-side features $\mathbf{f}_u$ (profiles, context, last 100 clicked items) and ad-side features $\mathbf{f}_a$ (metadata and real-time bid signals, Table~\ref{tab:bid_feature}). Categorical (ID) features are mapped to dense vectors by embedding lookup and numerical features are bucketized first, so that a single embedding space encodes heterogeneous feature types.

\subsection{Learning‐To‐Rank Paradigm}\label{sec:ltr}
Regressing eCPM directly is hard under the retrieval stage's capacity and latency budget. Following COPR~\cite{Zhao_2023} in pre-ranking, we adopt a pairwise learning-to-rank (LTR) formulation that targets the correct ordering of ads by commercial value, over the pairwise training set
\begin{align}
\mathcal{D}_{\mathrm{pair}}
&= \Bigl\{(u,\,a^{+}\in\mathcal{I}_u,\;a^{-}\in\mathcal{R}_u)\Bigr\} \notag \\
&\cup
\Bigl\{(u,\,a^{+}\in\mathcal{I}_u\cup\mathcal{R}_u,\;a^{-}\in\mathcal{A}\setminus(\mathcal{I}_u\cup\mathcal{R}_u))\Bigr\},
\end{align}
whose {\bf hard pairs} draw $a^{+}$ from the impression set $\mathcal{I}_u$ and the negative from the ranking set $\mathcal{R}_u$, and whose {\bf easy pairs} draw $a^{+}$ from $\mathcal{I}_u\cup\mathcal{R}_u$ and the negative from the remainder $\mathcal{A}\setminus(\mathcal{I}_u\cup\mathcal{R}_u)$. We train $f_{\mathrm{eCPM}}(u,a)$ to satisfy $f(u,a^{+})>f(u,a^{-})$ via a learn-to-rank loss~\cite{burges2005learning}:
\begin{equation}
\mathcal{L}_{\text{LTR}} = \mathbb{E}_{(u, a^{+}, a^{-}) \sim D_{\text{pair}}}  \left[ \log \left(1+e^{-[f_{\text{eCPM}}(u, a^{+})-f_{\text{eCPM}}(u, a^{-})]}\right) \right].
\end{equation}
The formulation is adapted from pre-ranking practice~\cite{Zhao_2023} and is not itself a contribution of this work, since it was already deployed in our production retriever before BAR (Section~\ref{sec:cumulative}); it is nonetheless the foundation on which the bidding-aware components below build.

\subsection{Bidding-Aware Modeling}\label{sec:bid_aware_retrieval}

\paragraph{Bidding-Aware Objective}

By the incentive compatibility principle~\cite{myerson1979incentive, GSP}, the predicted eCPM should vary monotonically with key bidding features such as bid amount and budget.
We instill this inductive bias as a soft constraint through data augmentation. For a training instance $(u, a)$ we build a synthetic counterpart $(u, \tilde{a})$ by stochastically perturbing one salient bid feature of $a$: budget\_left\_ratio is resampled uniformly from $[0,1]$, and bid\_constraint is multiplied by a ratio drawn uniformly from $[0,2]$. The model is then penalized whenever its two predictions violate the direction the perturbation implies. The resulting monotonicity loss is again pairwise logistic:
\begin{align}
\mathcal{L}_{\text{BAO}} = \mathbb{E}_{(u, a) \sim \mathcal{D}} \left[ \log \left(1 + e^{-\mathbb{I} \cdot [f_{\text{eCPM}}(u, \tilde{a}) - f_{\text{eCPM}}(u, a)]} \right) \right].
\end{align}

We adopt simple synthetic perturbations rather than coupling to a specific auction simulator, which keeps the framework independent of continuously evolving downstream auto-bidding strategies. Enforcing only a soft directional consistency between bid features and predicted scores, the heuristic acts as a regularizer that stays robust as bidding behavior shifts.

\paragraph{Distillation Auxiliary Objective}  
To enrich the retrieval model's representations with fine‐grained signals from the downstream ranking stage, we add two auxiliary distillation tasks, for pCTR and pBid prediction, that share the backbone with the main retrieval network:

\begin{align}
\mathcal{L}_{\mathrm{DAO}}
&=\mathcal{L}_{\mathrm{pCTR}}+\mathcal{L}_{\mathrm{pBid}}\,,\\
\mathcal{L}_{\text{pCTR}}
&=\mathbb{E}_{(u,a,\mathrm{pCTR})\sim\mathcal{D}}
\Bigl[\bigl(f_{\text{pCTR}}(u, a)-\mathrm{pCTR}\bigr)^{2}\Bigr]\,,\\
\mathcal{L}_{\text{pBid}}
&=\mathbb{E}_{(u,a,\mathrm{pBid})\sim\mathcal{D}}
\Bigl[\bigl(f_{\text{pBid}}(u, a)-\tfrac{\mathrm{pBid}}{\mathrm{aBid}}\bigr)^{2}\Bigr]\,,
\end{align}
where \(f_{\text{pCTR}}(u, a)\) and \(f_{\text{pBid}}(u, a)\) are the model's scalar outputs for click‐through rate and bid prediction. Normalizing the ground‐truth bid by the ad's default bid value \(\mathrm{aBid}\) (Table~\ref{tab:bid_feature}) stabilizes training across the wide dynamic range of real‐time bids.
The composite objective combines the primary LTR loss with the two auxiliary losses:
\begin{align}
\mathcal{L}_{\text{total}} = \mathcal{L}_{\text{LTR}} + \lambda_1 \mathcal{L}_{\text{BAO}} + \lambda_2 \mathcal{L}_{\text{DAO}} 
\label{eq:total_loss}
\end{align}
with $\lambda_1, \lambda_2 > 0$ weighting the auxiliary terms.

\subsection{Task-Attentive Refinement Module}\label{sec:model_architecture}

\begin{figure}[!tb]
  \centering
  \includegraphics[width=0.9\linewidth]{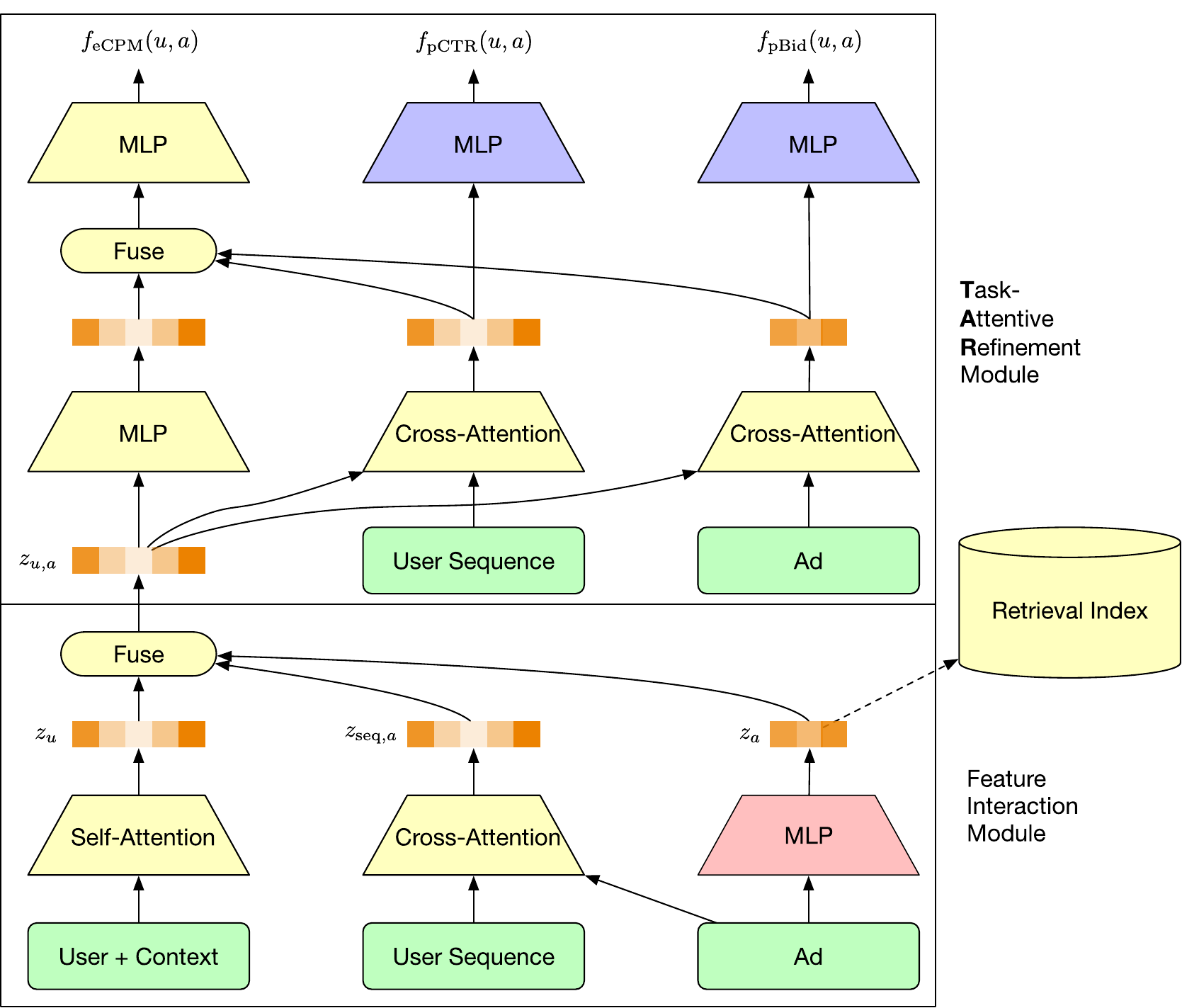}
  \caption{Architecture of BAR framework. Yellow, Red, Purple correspond to online, near-line, and offline modules.}
  \label{fig:network_architecture_aug}
  \Description{A schematic diagram showing the architecture of the BAR framework.}
\end{figure}

Our baseline model~\cite{Chen2022} captures fine-grained user-ad signals by cross-attention and retrieves top-$k$ candidates through an HNSW index~\cite{malkov2018efficient, douze2024faiss}. Its lower ``Feature Interaction Module'' in Figure~\ref{fig:network_architecture_aug} has three parallel branches: a user network producing a high-level user embedding $z_u$ from profiles and context; a cross-attention network over the user behavior sequence and the ad features producing a contextually aware interaction embedding $z_{\text{seq}, a}$; and an ad network compressing ad features into a succinct $z_a$ for indexing. Their concatenation $z_{u, a}$ feeds MLP task heads that output $f_{\text{eCPM}}(u, a)$, $f_{\text{pCTR}}(u, a)$, and $f_{\text{pBid}}(u, a)$.

This fusion is representationally imbalanced. In our production environment $z_u$ (dim $>$ 1000) dwarfs $z_{\text{seq}, a}$ (dim = 64) and $z_a$ (dim = 64, constrained by ANN indexing~\cite{Li2016ApproximateNN}), so it risks overshadowing the nuanced signals of the interaction and ad branches, above all the real-time bidding information carried by $z_a$. Widening $z_a$ is impractical, as the added dimensionality invites the curse of dimensionality and degrades indexing accuracy, while shrinking $z_u$ only creates an information bottleneck that dilutes sparse but critical signals.

To resolve this while aligning the architecture with our multi-task objectives, we place a \textbf{T}ask-\textbf{A}ttentive \textbf{R}efinement (\textbf{TAR}) module atop $z_{u, a}$, with three specialized branches:
\begin{itemize}[leftmargin=*]
    \item \textbf{pCTR Head}: this branch models user interest by cross-attending $z_{u, a}$ to the user's historical behavior sequence, and passes the result through an MLP that produces $f_{\text{pCTR}}(u, a)$.
    \item \textbf{pBid Head}: this branch sharpens sensitivity to commercial signals by cross-attending $z_{u, a}$ to the raw ad features, which lets the model attend explicitly to dynamic bidding attributes before an MLP computes $f_{\text{pBid}}(u, a)$.
    \item \textbf{eCPM Head}: this branch combines the two. It fuses the penultimate representations of the pCTR and pBid heads with a linear projection of $z_{u,a}$ in an MLP that emits the holistic score $f_{\text{eCPM}}(u, a)$.
\end{itemize}
TAR therefore grants user interest and bidding dynamics dedicated representational capacity before combining them. Online, only $f_{\text{eCPM}}(u, a)$ drives top-$k$ retrieval; the auxiliary outputs serve training alone.

\subsection{System Implementation} \label{sec:system_implementation}

\begin{figure}[!tb]
  \centering
  \includegraphics[width=0.9\linewidth]{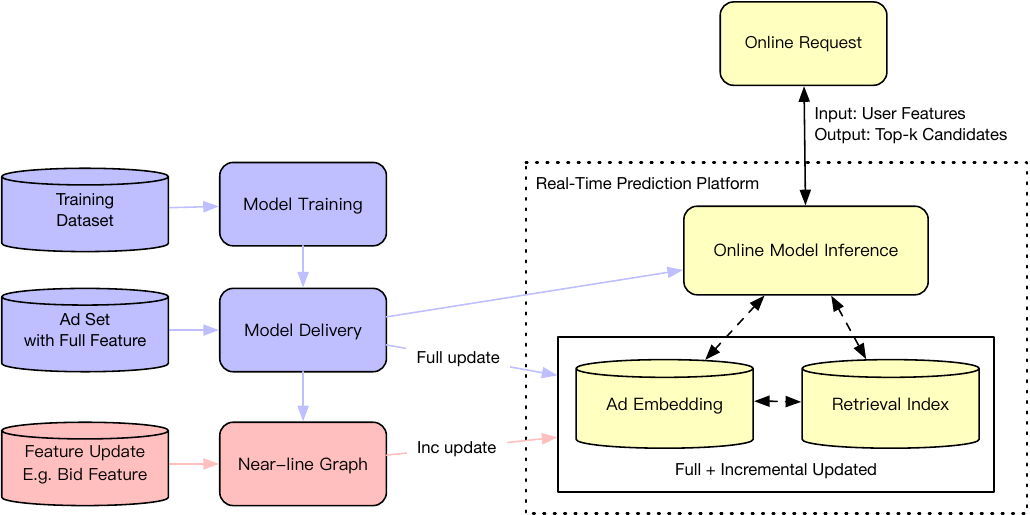}
  \caption{System architecture of the \textbf{BAR} Framework. The distinct roles of the Offline (purple), Near-Line (red), and Online (yellow) components are illustrated.}
  \Description{A schematic diagram showing the architecture of the BAR framework, including modules for real-time bid-aware retrieval and asynchronous near-line inference.}
  \label{fig:system_architecture}
\end{figure}

Pre-computed ad representations are static, which puts the ad embeddings, refreshed only infrequently, at odds with attributes as volatile as bid prices, and blunts the model's ability to track market signals. Our Asynchronous Near-Line Inference framework (Figure~\ref{fig:system_architecture}) removes that gap with three interconnected components.

\paragraph{Offline} Besides training, this component delivers the model as two decoupled computational graphs (cf.\ Figure~\ref{fig:network_architecture_aug}): a \textit{User-Side Graph} holding the user network and the user-ad interaction components, and an \textit{Ad-Side Graph} holding only the ad network that generates ad embeddings. At deployment we run the Ad-Side Graph over the whole corpus in one batch and build the HNSW index over the resulting ad embeddings, with $M{=}64$ neighbors per node and \texttt{efConstruction} left at the FAISS default~\cite{douze2024faiss}. The index and the User-Side Graph are then shipped to the serving fleet, and the Ad-Side Graph is deployed a second time, to a dedicated near-line service. This separation is what isolates online query processing from the embedding update path.

\paragraph{Online} Engineered for low latency and high throughput, this service answers each request with the User-Side Graph and the HNSW index, returning the top-$k$ ads. Retrieval follows the beam-retrieval of NANN~\cite{Chen2022}: the graph is traversed top-down, each layer expanding a dynamic candidate list for a fixed number of steps, with every visited ad scored by the eCPM head rather than by a metric distance; the fixed step count is what bounds the tail latency. The index is dynamic: it absorbs incremental updates pushed from the near-line system, so retrieval always runs against the freshest available ad representations.

\paragraph{Near-Line} The cornerstone of BAR's responsiveness is an event-driven pipeline that keeps ad embeddings and the HNSW index fresh. Triggered by ad feature modifications such as bid constraint adjustments and budget changes, which arrive at about $10^{4}$ QPS, the service invokes the Ad-Side Graph to recompute the embedding and HNSW neighbors of the affected ad alone, and then propagates both asynchronously to the online fleet. Each update payload is correspondingly small, a 64--256\,D embedding plus a 32--128\,D neighbor list, and stays below 1\,MB.

These writes mutate a live index under a read-heavy, highly concurrent access pattern, in which read QPS in the millions dwarfs the write traffic, so locking is kept at the granularity of a single ad entry: a readers-writer spinlock per entry admits concurrent readers, and a writer holds it only long enough to swap one embedding and its neighbor list. A write thus completes within 0.5\,ms at P99.9 without blocking any user request, and ad representations converge within seconds of an advertiser action.

\section{Experiments}
\subsection{Setup}
\paragraph{Architectures}
We validate BAR on the two retrieval architectures that bracket the design space in production. The first is the model-based retriever \textbf{MBR}~\cite{Chen2022}, which scores user-ad pairs by cross-attention and searches a graph-based HNSW index; it receives all three of our components, BAO, DAO and TAR. The second is the embedding-based retriever \textbf{EBR}, instantiated as SASRec~\cite{Kang2018SelfAttentiveSR}, which receives BAO and DAO only. TAR is inapplicable there by construction: it refines a fused user-ad representation, and a two-tower model never forms one, since its user and ad sides meet only at the final inner product.

\paragraph{Dataset}
Because no publicly available dataset exposes bid features, all experiments are conducted on production data sampled from Alibaba's display advertising platform. The sample spans the 8 days from June 28 to July 5, 2025, and comprises 3 billion sessions, each carrying 5 impression ads together with 15 ads sampled from those that reached the ranking stage. We train on the first 7 days and evaluate on the last one, so that evaluation always looks forward in time. Across the sample, 200 million unique users interact with 50 million unique ads, and each session carries 90 user profile, 10 context, 100 behavior sequence and 80 ad profile features.

\paragraph{Training}
We train every model for a single epoch with AdamW on NVIDIA H20 GPUs, in TensorFlow on XDL~\cite{jiang2019xdl}, embedding each feature into 32 dimensions. MBR uses a learning rate of 4.8e-3, a weight decay of 1e-4 and a batch size of 30,000, and consumes roughly 1{,}500 GPU hours at $\lambda_1{=}0.5$ and $\lambda_2{=}1.0$. EBR follows the hyperparameters of SASRec~\cite{Kang2018SelfAttentiveSR}, consumes roughly 750 GPU hours, and uses $\lambda_1{=}0.1$ and $\lambda_2{=}0.5$.

\subsection{Offline Results}
\subsubsection{Effectiveness of the Bidding-Aware Objective}

\begin{table}[!tb]
\centering
\caption{The effectiveness of bidding-aware objective and bid features under different model architectures.}
\begin{tabular}{cccc}
\toprule
\textbf{Model} & \textbf{Ablation} & \textbf{Recall@2000} & \textbf{PairAcc} \\
\midrule
\multirow{3}{*}{MBR} & BAR & \textbf{54.9\%} & \textbf{96.1\%} \\
& w/o BAO & 52.6\% & 80.2\%  \\
& w/o BF & 50.9\% & 50.0\% \\
\midrule
\multirow{3}{*}{EBR} & BAR & \textbf{38.7\%} & \textbf{94.3\%} \\
& w/o BAO & 36.5\% & 65.6\% \\
& w/o BF & 36.4\% & 50.0\% \\
\bottomrule
\end{tabular}
\label{tab:bid_aware_rst}
\end{table} 

We compare the full framework against two ablations. \textbf{w/o BAO} drops $\mathcal{L}_{\text{BAO}}$ from Eq.~\ref{eq:total_loss} while keeping the bid features in the input, which isolates the monotonicity constraint from the features it constrains. \textbf{w/o BF} additionally removes every explicit bid feature (Table~\ref{tab:bid_feature}) from the input; BAO has nothing left to perturb and is therefore omitted as well, leaving a fully bid-agnostic retriever.

Following \cite{Chen2022}, we use the impression $\text{Recall}@M$ metric to evaluate retrieval quality. For request $u$, let $\mathcal{P}_u$ ($|\mathcal{P}_u|{=}M$) and $\mathcal{G}_u$ denote the retrieved ad set and the ground-truth impression set, respectively:
\begin{equation}
\text{Recall}@M = \mathbb{E}_{u \sim U} \left[ \frac{|\mathcal{P}_u \cap \mathcal{G}_u|}{|\mathcal{G}_u|} \right],
\end{equation}
with $M{=}2000$ in our experiments. To directly quantify monotonic compliance under bid-feature perturbations, we additionally introduce Pairwise Accuracy (PairAcc), the proportion of samples where the predicted score changes in the correct direction:
\begin{align}
\text{PairAcc} &= \mathbb{E}_{(u, a)} \left[ \text{mono}(\mathbb{I}, u, a, \tilde{a}) \right], \\
\text{mono}(\mathbb{I}, u, a, \tilde{a}) &=
\begin{cases}
1, & \mathbb{I}{=}1 \text{ and } f_{\text{eCPM}}(u, \tilde{a}) > f_{\text{eCPM}}(u, a) \\
1, & \mathbb{I}{=}{-1} \text{ and } f_{\text{eCPM}}(u, \tilde{a}) < f_{\text{eCPM}}(u, a) \\
0, & \text{otherwise}.
\end{cases}
\end{align}

Both metrics are bounded by what they can observe: Recall is scored against the exposure of the incumbent system, and PairAcc against our own monotonicity prior, so the online results of Section~\ref{sec:online}, where candidates are genuinely served, are what ultimately test them. Nor do we verify the perturbation magnitudes of Section~\ref{sec:bid_aware_retrieval} against logged bid-change distributions; a model tuned to them might use its capacity more efficiently, but since the downstream bidding strategies change without the retriever being retrained, we prefer the robustness.
Table~\ref{tab:bid_aware_rst} confirms BAO's role: ablating it costs 2.3 points of Recall@2000 on MBR (54.9\% to 52.6\%) and 2.2 on EBR (38.7\% to 36.5\%), so the effect carries across architectures. PairAcc adds nuance: without BAO the model already picks up an implicit bid-outcome correlation (well above 50\%), and our objective turns that implicit signal into an explicit auction-aware constraint, which acts as a powerful regularizer and lifts Recall along with PairAcc.

\subsubsection{Effectiveness of Distillation Auxiliary Objective} 
Ablating the Distillation Auxiliary Objective (w/o DAO) degrades both architectures (Table~\ref{tab:mul_task_rst}): Recall@2000 falls from 54.9\% to 54.1\% on MBR and from 38.7\% to 37.9\% on EBR. Supervising the two constituents of eCPM separately tells the shared backbone how the click probability and the bid each contribute, not merely which ad of a pair ranks higher, and the primary task inherits the better representation.

\begin{table}[!tb]
\centering
\caption{The effectiveness of distillation auxiliary objective under different model architectures.}
\begin{tabular}{ccc}
\toprule
\textbf{Model} &\textbf{Ablation} & \textbf{Recall@2000} \\
\midrule
\multirow{2}{*}{MBR} & BAR &  \textbf{54.9\%} \\
& w/o DAO & 54.1\%   \\
\midrule
\multirow{2}{*}{EBR}& BAR  & \textbf{38.7\%} \\
& w/o DAO & 37.9\%  \\
\bottomrule
\end{tabular}
\label{tab:mul_task_rst}
\end{table} 

\subsubsection{Sensitivity Analysis of Loss Weights}

\begin{figure}[!tb]
  \centering
  \begin{subfigure}{0.51\linewidth}
    \centering
    \includegraphics[width=\linewidth]{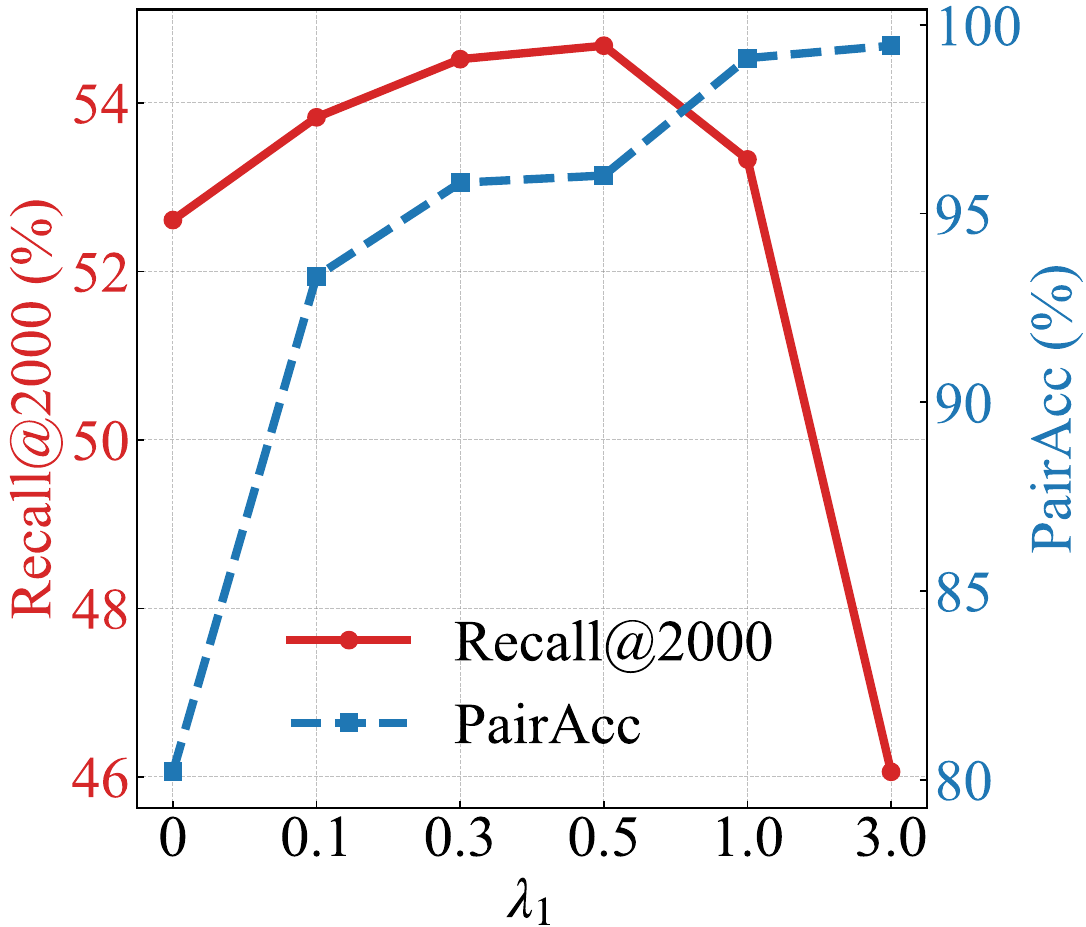}
    \caption{hyperparameter $\lambda_1$}
    \label{fig:lambda1_result}
  \end{subfigure}
  \hspace{0.01\linewidth}
  \begin{subfigure}{0.44\linewidth}
    \centering
    \includegraphics[width=\linewidth]{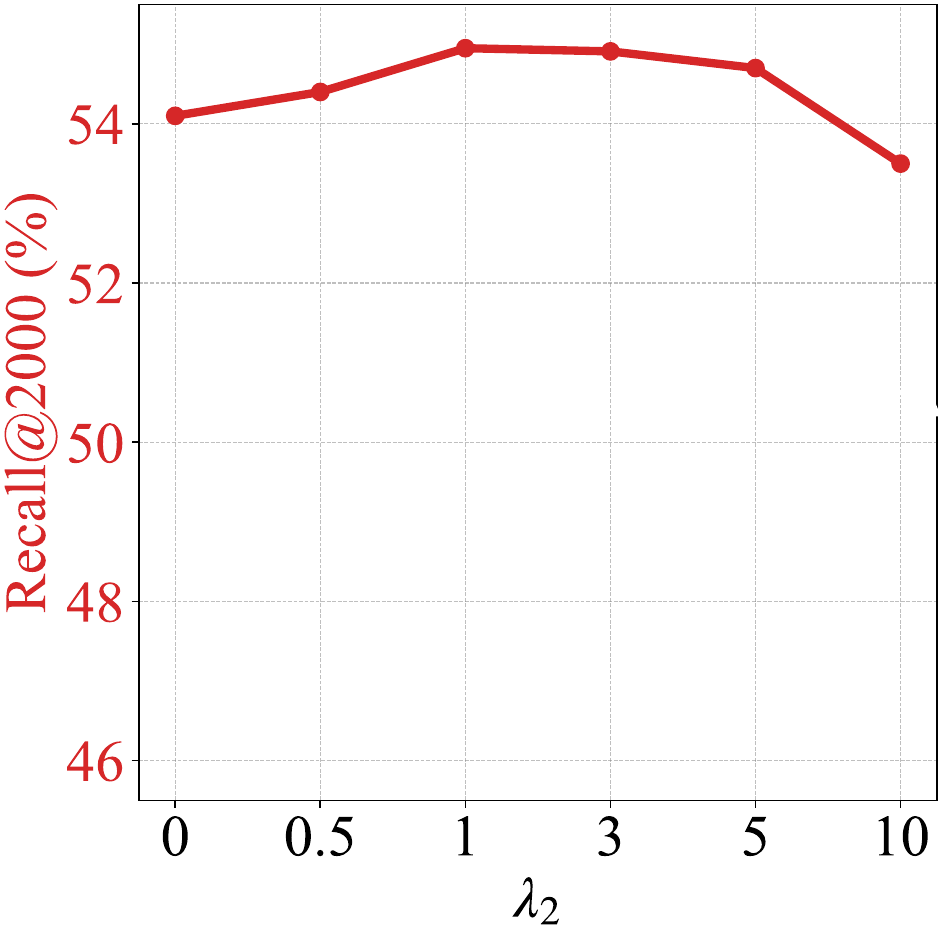}
    \caption{hyperparameter $\lambda_2$}
    \label{fig:lambda2_result}
  \end{subfigure}
  \caption{Sensitivity analysis of loss weights $\lambda_1$ (BAO) and $\lambda_2$ (DAO) on the MBR architecture.}
  \Description{Two line charts. The left chart shows that as the BAO weight increases, PairAcc rises monotonically while Recall@2000 follows a concave curve peaking around 0.3 to 0.5. The right chart shows Recall@2000 remaining nearly flat across DAO weights from 0 to 5 and dropping sharply at a weight of 10.}
  \label{fig:lambda_result}
\end{figure}

On the MBR model, the bidding-aware weight $\lambda_1$ shows the classic multi-task trade-off (Figure~\ref{fig:lambda_result}(\subref{fig:lambda1_result})): PairAcc improves monotonically with $\lambda_1$ while Recall@2000 is concave, i.e.\ enforcing economic coherence helps until it crowds out other patterns. Performance is nonetheless robust throughout $[0.3, 0.5]$, which keeps tuning simple, and EBR behaves similarly.
The model is far less sensitive to the distillation weight $\lambda_2$ (Figure~\ref{fig:lambda_result}(\subref{fig:lambda2_result})): performance stays near-optimal across $[0, 5]$, peaks at $\lambda_2{=}1$, and degrades only at exceptionally large values such as $\lambda_2{=}10$.

\subsubsection{Effectiveness of Task-Attentive Refinement Module}
\begin{table}[!tb]
\centering
\caption{Comparative results on the effectiveness and computational efficiency of TAR Module.}
\begin{tabular}{lccc}
\toprule
\textbf{Model} & \textbf{Recall@2000} & \textbf{Recall-all@2000} & \textbf{GFlops} \\
\midrule
Baseline  & 51.5\% & 64.3\%  &  1.0x  \\
+ IncDim & 46.2\% & 66.8\% & 4x \\
+ TAR & \textbf{54.9\%} & \textbf{68.2\%} & \textbf{1.2x} \\
\bottomrule
\end{tabular}
\label{tab:tar_rst}
\end{table}

We test on the MBR architecture whether TAR really addresses the representational imbalance of Section~\ref{sec:model_architecture}, comparing it against two variants:
\begin{itemize}[leftmargin=*]
    \item \textbf{Baseline}~\cite{Chen2022}: predicts the final eCPM directly from the concatenated $z_{u,a}$ via three separate MLPs dedicated to eCPM, pCTR, and pBid.
    \item \textbf{+ IncDim}: a brute-force fix that expands the interaction embedding $z_{\text{seq}, a}$ and ad embedding $z_a$ from 64 to 1024 to align with $z_u$.
\end{itemize}
Since ``+ IncDim'' is not indexable by conventional ANN at 1024 dimensions, where the curse of dimensionality degrades indexing accuracy~\cite{Li2016ApproximateNN}, we additionally report Recall-all@2000 by full-corpus scoring. Table~\ref{tab:tar_rst} shows ``+ IncDim'' improving Recall-all@2000 over the baseline at a prohibitive 4$\times$ FLOPs cost, whereas ``+ TAR'' attains the best 68.2\% for only 1.2$\times$, harmonizing user-interest and commercial signals at a deployment-feasible cost.
The three heads also learn distinct representations. In the Spearman correlations~\cite{schober2018correlation} of Table~\ref{tab:tar_head_correlation}, the pCTR head correlates strongly with pCTR (0.672) and negatively with the bid ($-0.325$), the pBid head shows the mirror image (0.702 with the bid and $-0.327$ with pCTR), and the eCPM head sits between them, correlating moderately with both.

\begin{table}[!tb]
\centering
\caption{Spearman correlation between TAR head outputs and targets.}
\begin{tabular}{lccc}
\toprule
\textbf{Target} & \textbf{eCPM Head} & \textbf{pCTR Head} & \textbf{pBid Head} \\
\midrule
eCPM & \bf{36.9\%} & 22.5\% & 30.6\% \\
pCTR & 15.8\% & \bf{67.2\%} & -32.7\% \\
Bid & 26.7\% & -32.5\% & \bf{70.2\%} \\
\bottomrule
\end{tabular}
\label{tab:tar_head_correlation}
\end{table}

\subsubsection{Cumulative Contribution of Each Component}\label{sec:cumulative}
To delineate the boundary of our contributions, we build BAR up one component at a time on the MBR architecture, starting from a retriever trained on the impression set $\mathcal{I}_u$ only (Table~\ref{tab:cumulative}).
Two points stand out. First, the pairwise LTR paradigm accounts for the single largest jump ($+10.3$\%), but it was already deployed in our production system before this work and is therefore part of the online baseline of Section~\ref{sec:online} rather than a contribution of BAR.
Second, exposing raw bid features to the retriever is almost useless on its own ($+0.1$\%); the gain materializes only once $\mathcal{L}_{\text{BAO}}$ imposes economic monotonicity on them ($+2.7$\%), which rules out the reading that BAR merely benefits from having bid features in the input. DAO and TAR then add a further $+0.6$\% and $+3.4$\%, the latter being the largest increment contributed by any single component we introduce.

\begin{table}[!tb]
  \centering
  \caption{Cumulative contribution of each component on the MBR architecture. $\Delta$ is the increment over the preceding row.}
  \label{tab:cumulative}
  \begin{tabular}{lcc}
    \toprule
    \textbf{Configuration} & \textbf{Recall@2000} & \textbf{$\Delta$} \\
    \midrule
    Base ($\mathcal{I}_u$ only) & 37.8\% & --- \\
    + $\mathcal{R}_u$ \& LTR (production baseline) & 48.1\% & +10.3\% \\
    + Bid Features & 48.2\% & +0.1\% \\
    + BAO & 50.9\% & +2.7\% \\
    + DAO & 51.5\% & +0.6\% \\
    + TAR (full BAR) & \textbf{54.9\%} & +3.4\% \\
    \bottomrule
  \end{tabular}
\end{table}

Since no public retrieval baseline ingests dense bid signals, these rows double as our points of comparison: ``+ Bid Features'' is a bid-feature-augmented retriever without a monotonicity constraint, and ``+ DAO'' adds the standard distillation of downstream ranking scores. Replacing $\mathcal{L}_{\text{LTR}}$ with a sampled softmax over impressions, i.e.\ direct value prediction, yields 42.3\%.

\subsection{Online Results}\label{sec:online}
We ran a month-long online A/B test in the Alibaba display advertising system, which spans multiple business scenarios and serves billions of users, randomly partitioning 10\% of online traffic into two groups:

\textbf{Baseline}: the incumbent production system, whose retrieval is multi-channel. Its primary channel, denoted the `ad value' channel, runs the same MBR backbone as BAR, that is, cross-attention scoring over an HNSW index~\cite{Chen2022} trained with the identical pairwise LTR objective on the identical data. What it lacks are our contributions: it has neither Bidding-Aware Modeling nor Task-Attentive Refinement, it consumes no bid features, and its ad embeddings are pre-computed in daily full batches instead of being refreshed by Asynchronous Near-Line Inference. It thus corresponds exactly to the ``+ $\mathcal{R}_u$ \& LTR'' configuration of Table~\ref{tab:cumulative}, which makes the online comparison an ablation of bid-awareness rather than of model capacity.

\textbf{BAR}: the baseline model is replaced with our framework, all other settings identical. Beyond the 10\% main test reported below, BAR was validated over the same month-long horizon by a 50\% scaled comparison and a 5\% reverse (holdout) experiment, both reproducing consistent and stable gains.

\paragraph{Serving cost of the trade-off}
Under a 40\,ms latency constraint, BAR sustains 500 QPS against 600 QPS for the baseline. This 17\% throughput reduction is attributable entirely to the 20\% GFlops increase of the TAR module: measured in isolation, BAR without near-line inference shows the same 1.2$\times$ GFlops and the same 500 QPS, since embedding updates never touch the online critical path.
At our scale the loss is a real infrastructure cost, but narrowly scoped: it applies only to the bidding-aware channel, which consumes 10\% of the retrieval candidate quota while contributing over half of the final impressions, so the extra capacity is a small fraction of the retrieval fleet and is outweighed by the revenue gain below.

\subsubsection{Online Metrics on Platform Perspective}
Alongside total Platform Revenue we track three ratios, defined as follows: Revenue Per Mille (RPM) assesses ad monetization, Click-Through Rate (CTR) captures the user experience, and Return On Investment (ROI) measures the advertiser's experience.
\begin{align}
\text{RPM} &= \frac{\text{Platform Revenue}}{\text{\# of impressions}} \times 1000, \\
\text{CTR} &= \frac{\text{\# of clicks}}{\text{\# of impressions}}, \\
\text{ROI} &= \frac{\text{GMV of advertisers}}{\text{Platform Revenue}}.
\end{align}

BAR significantly enhances platform performance, yielding {\bf a 4.32\% increase in Platform Revenue} (95\% CI: [3.93\%, 4.71\%], $p<0.0001$) and a {\bf 3.78\% increase in RPM} (95\% CI: [3.37\%, 4.15\%], $p<0.0001$), without compromising user experience or advertiser effectiveness: CTR (+0.31\%) and ROI (+0.01\%) both stay stable.
We stress that this revenue gain is the \emph{joint} effect of the three components: a month-long, per-component traffic split was not affordable in production, so BAR is evaluated online only against the bid-agnostic baseline. The attribution of the individual components rests on offline evidence (Table~\ref{tab:cumulative}) and, for Asynchronous Near-Line Inference, on the responsiveness metrics below, which cannot improve at all while ad embeddings remain static.

\subsubsection{Responsiveness to Advertiser Operations}

\begin{figure}[!tb]
  \centering
  \includegraphics[width=\linewidth]{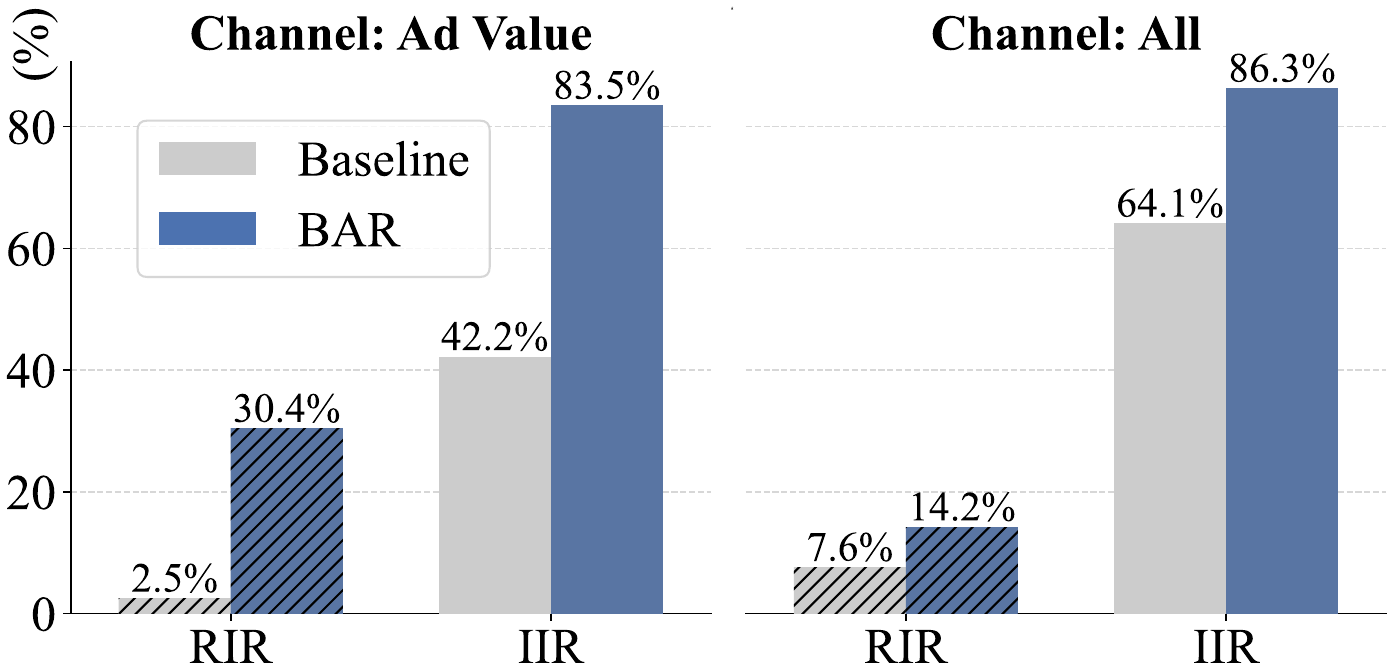}
  \caption{Impact of BAR on System Responsiveness: Comparison of RIR and IIR metrics}
  \Description{Grouped bar charts comparing the baseline and BAR on the Retrieval Improved Ratio and the Impression Improved Ratio, reported both for the ad-value retrieval channel and across all retrieval channels; BAR is substantially higher on every group.}
  \label{fig:advertiser_rst}
\end{figure}

We quantify responsiveness to advertiser activity with the Impression Improved Ratio (IIR) and the Retrieval Improved Ratio (RIR):
\begin{align}
    \text{IIR} &= \mathbb{E}_{a \sim A^{+}} \left[ \frac{NI_{a, {h^+}}}{NI_{a, {h^-}}} - 1 \right], \\
  \text{RIR} &= \mathbb{E}_{a \sim A^{+}} \left[ \frac{NR_{a, {h^+}}}{NR_{a, {h^-}}} - 1 \right]
\end{align}
where $A^+$ is the set of ads with positive operations, and $NI_{a, h^-}$, $NI_{a, h^+}$ count the impressions of ad $a$ in the $h$-hour windows preceding and following the operation time $t$, i.e.\ in $[t-h, t]$ and $[t, t+h]$; $NR_{a, h^-}$ and $NR_{a, h^+}$ count the times $a$ was retrieved among the top-$k$ results in those same windows. Both measure how well the system surfaces an ad after a positive advertiser action such as a bid increase ($h=3$ in our experiments).

In the ``Ad Value'' channel (Figure~\ref{fig:advertiser_rst}), RIR escalates from a mere 2.5\%, a baseline signifying near-complete insensitivity to advertiser actions, to 30.4\%. This over twelve-fold increase reflects a substantially stronger capacity to retrieve candidates pertinent to an advertiser's bidding adjustments, and it carries IIR from 42.2\% to 83.5\% in the same channel. Asynchronous Near-Line Inference is what enables these gains: updating ad embeddings as advertisers act overcomes the staleness of conventional pre-computation and lets the model follow market dynamics in near-real-time.
The effect propagates system-wide. Measured across all retrieval channels, RIR nearly doubles from 7.6\% to 14.2\% and IIR rises from 64.1\% to 86.3\%, i.e.\ gains of 6.6 and 22.2 percentage points.

\subsubsection{Analysis by Bid Type}
The gains hold across all major bidding strategies. Offline, Recall improves by 6.5\% for OCPC, 6.2\% for Max Return and 5.9\% for Multi-constrained bidding. Online, no strategy loses meaningful ground in cost share: OCPC gains 1.1 points, while Max Return and Multi-constrained bidding give up 0.3 and 0.8. BAR therefore improves every strategy rather than reallocating budget from one to another.

\subsubsection{Fairness and Diversity Analysis}\label{sec:fairness}
BAR promotes a healthier ecosystem by letting long-tail ads and advertisers earn visibility through active operation instead of being buried by historical CTR bias: the impression and cost share of long-tail ads (fewer than 1k user clicks per day) rose by 1.6\% and 0.7\%, so the altered candidate distribution is the more inclusive one.
This responsiveness does not collapse into a bid-driven echo chamber either, because retrieval stays multi-channel: parallel channels built on multi-modal and LLM-based representations keep injecting candidates no bid signal would select, and we observed no drift over the month-long horizon.

\section{Conclusion}
This paper introduces BAR (Bidding-Aware Retrieval) to improve multi-stage consistency between retrieval and the stages that follow it. Three components carry it: Bidding-Aware Modeling, whose monotonicity constraint and distillation losses yield economically coherent representations; Task-Attentive Refinement, which disentangles user interest from commercial value; and Asynchronous Near-Line Inference, which adapts to bid changes in near-real time. Full deployment on Alibaba's display advertising platform validates the design under industrial-scale workloads: $+4.32$\% platform revenue, stronger advertiser control ($+6.6$ and $+22.2$ percentage points in RIR and IIR), and a maintained user experience ($+0.31$\% CTR).

\begin{acks}
We sincerely appreciate Zhengxiong Zhou and Huimin Yi for implementing the key components of the training framework, Junchen Dong, Wenchao Wang, Zhengyu Liu and Yunlong Xu for the inference engine, and Jiaqi An and Fei Wang for the data infrastructure.
\end{acks}

\clearpage
\section*{GenAI Usage Disclosure}
In accordance with the ACM Policy on the use of AI, the authors disclose that generative AI tools were used solely to assist with manuscript writing, including language polishing, phrasing suggestions, and formatting. All AI-assisted content was subsequently reviewed, verified, and edited by the human authors, who take full responsibility for the final text. No generative AI tools were used in the research process itself, including the design of the methodology, the development of the code, the collection or processing of data, or the analysis of experimental results.

%% the bibliography file
\bibliographystyle{ACM-Reference-Format}
\balance
\bibliography{sample-base}

\end{document}